\documentclass{article} % For LaTeX2e
\usepackage{iclr2026_conference,times}

\usepackage{amsmath,amsfonts,bm}

\def\eqref#1{equation~\ref{#1}}
\def\1{\bm{1}}

\DeclareMathAlphabet{\mathsfit}{\encodingdefault}{\sfdefault}{m}{sl}
\SetMathAlphabet{\mathsfit}{bold}{\encodingdefault}{\sfdefault}{bx}{n}

\usepackage{hyperref}
\usepackage{url}
\usepackage[pdftex]{graphicx}
\usepackage{cleveref}
\usepackage{enumitem}
\usepackage{tabularx}
\usepackage{booktabs}
\usepackage{float}
\usepackage{caption}
\usepackage{subcaption}
\usepackage{listings}
\usepackage{wrapfig}

\title{Path-Consistency with Prefix Enhancement for Efficient Inference in LLMs}

\iclrfinalcopy
\author{Jiace Zhu, Yuanzhe Huang, Yingtao Shen, An Zou \\
Shanghai Jiao Tong University \\
\texttt{\{zhujiace,2596143200,doctorcoal,an.zou\}@sjtu.edu.cn} \\
\AND
Jie Zhao \\
Microsoft Corporation \\
\texttt{zhaojie@microsoft.com}
}

\begin{document}

\maketitle

\begin{abstract}
To enhance the reasoning capabilities of large language models (LLMs), self-consistency has become a popular approach, combining multiple samplings with majority voting. However, current methods are computationally expensive and time-consuming due to the need for numerous samplings. To address this, this paper introduces \textit{path-consistency}, which leverages the confidence of earlier-generated answers to identify the most promising prefix and guide the generation of subsequent branches. By dynamically guiding the generation of subsequent branches based on this prefix, \textit{path-consistency} mitigates both the errors and redundancies from random or less useful sampling in self-consistency. This approach reduces errors and redundancies from random sampling, significantly accelerating inference by minimizing token consumption. Our extensive empirical results demonstrate that \textit{path-consistency} improves inference latency by up to 40.5\%, while maintaining task accuracy across various tasks, including mathematical reasoning, commonsense reasoning, and symbolic reasoning.
\end{abstract}

\section{Introduction}
% 引入cot和sc
 The range of tasks that large language models (LLMs) can accomplish is continuously expanding, as the scale and complexity of models continue to grow recently. However, this advancement has not yet endowed LLMs with sufficiently robust reasoning capabilities \cite{rae2021scaling, wu2016google, guo2018long, chen2022program}. To address this shortcoming and further extend the application scope of LLMs, \textit{Chain-of-Thought} (CoT) \cite{wei2022chain} prompting has emerged in response. CoT prompting uses reasoning problems and processes as input prompts to guide language models in generating reasoning paths and final answers, aiming to mimic the thought processes humans might use when solving mathematical or logical problems. This enables LLMs to be applied in an increasing number of specific scenarios, such as mathematical reasoning \cite{cobbe2021training, miao2020diverse} and logical reasoning \cite{geva2021did}. To better accomplish these complex tasks, there are several CoT-based optimization methods \cite{li2022advance, chen2022program}. One of the common and effective methods is \textit{self-consistency} \cite{wang2022self}, a technique involving multiple sampling and majority voting. In this approach, the model generates multiple reasoning paths for a given input, with the final decision based on the most frequently occurring output among the samples.
% sc的缺陷

\begin{figure*}[!tbhp]
\vspace{-5mm}
\centering
\begin{minipage}[t]{0.49\textwidth}
    \centering
    \includegraphics[width=\linewidth]{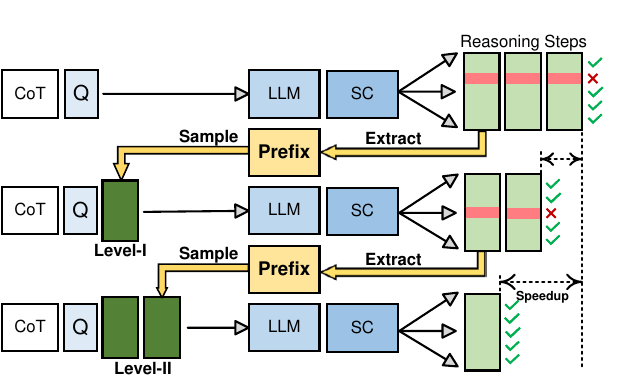}
    \caption{Path-consistency extracts prefixes from earlier generated inference paths to guide the inference of subsequent branches.}
    \label{fig:path-consistency-simple}
\end{minipage}
% \hspace{-0mm}
\begin{minipage}[t]{0.49\textwidth}
    \centering
    \includegraphics[width=\linewidth]{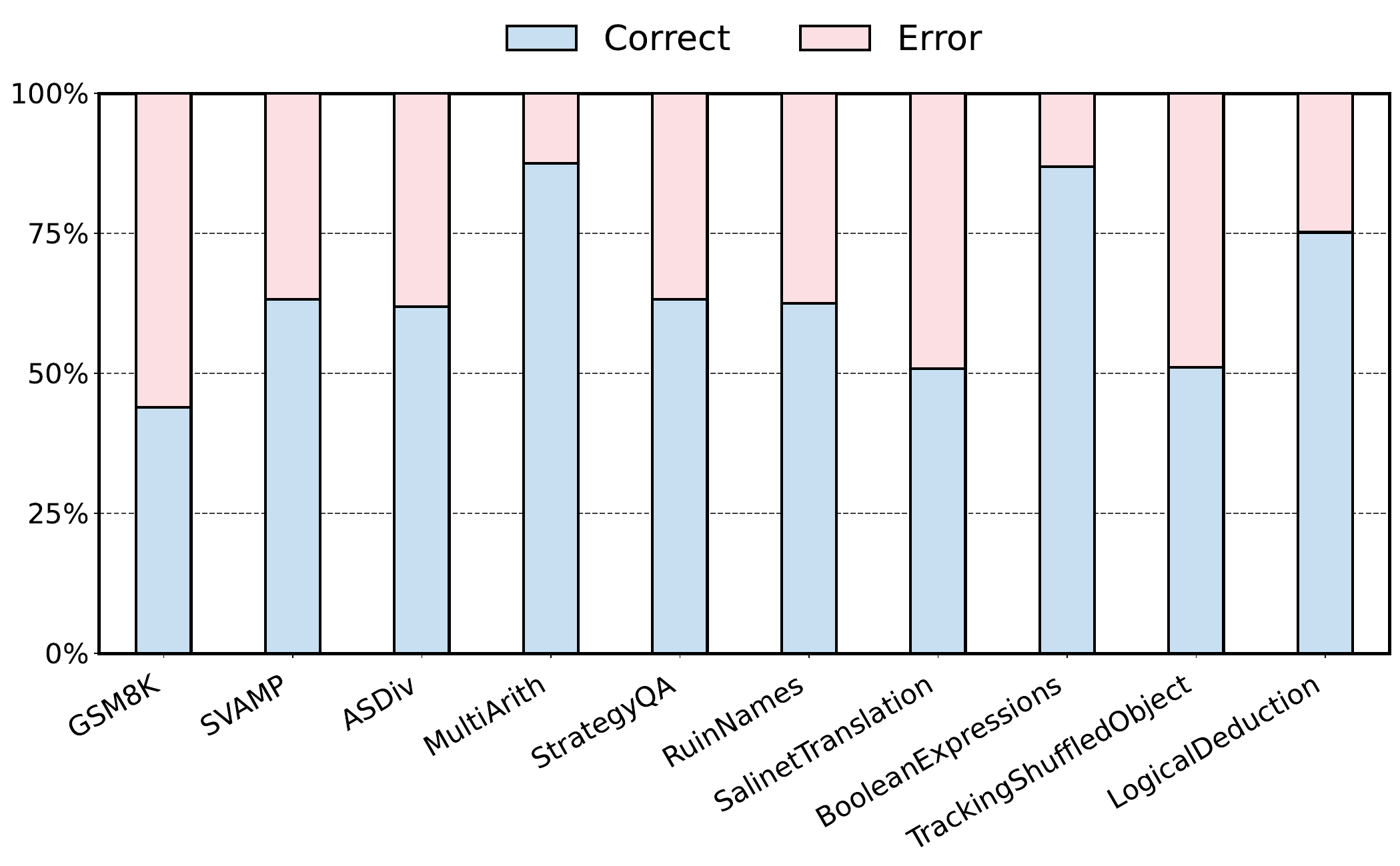}
    \caption{The proportion of tokens generated by self-consistency on correct or incorrect inference paths.}
    %  In most tasks, self-consistency wastes over 25\%, or even 50\%, of tokens on incorrect branches, resulting in significant additional costs.
    \label{fig:characterization}
\end{minipage}
\vspace{-5mm}
\end{figure*}

% \begin{figure}[!htp]
% \begin{center}
% \includegraphics[width=0.8\linewidth]{figures/Characterization.pdf}
% \end{center}
% \caption{The proportion of tokens generated by self-consistency on correct or incorrect inference paths. In most tasks, self-consistency wastes over 25\%, or even 50\%, of tokens on incorrect branches, resulting in significant additional costs.}
% \label{fig:characterization}
% \end{figure}

Self-consistency significantly enhances the reasoning capabilities of LLMs by sampling a large number of examples during inference. However, it has limitations in practical applications. The basic self-consistency technique frequently invokes the model to generate numerous reasoning paths when solving specific problems \cite{wang2022self}. As model size and task complexity increase, the time and computation cost of self-consistency increase sharply, making it a critical issue in practical applications. Motivated by this challenge, this paper proposes \textit{path-consistency}, which is designed to reduce the computation and time cost by leveraging intermediate information from earlier generation results to assist in subsequent generations. 

As illustrated in \cref{fig:path-consistency-simple}, the proposed dynamic inference method, path-consistency, continuously extracts appropriate reasoning steps from already generated reasoning paths to serve as ``prefixes''. These prefixes are then used to guide and accelerate the generation of subsequent reasoning steps. 

The proposed \textit{path-consistency} offers several key advantages: (1) It accelerates inference speed and reduces token consumption, while significantly preserving or even improving task accuracy. (2) The method requires no additional computation, fine-tuning, or training, ensuring the generation quality of the model remains intact. (3) It is model-agnostic, making it easy to deploy and apply to various models and tasks in practical scenarios. Furthermore, it integrates seamlessly with existing optimization methods, achieving even better acceleration performance.

We evaluated path-consistency using ten different datasets, resulting in an average acceleration of 28.7\% for mathematical reasoning tasks, 20.9\% for commonsense reasoning, and 20.3\% for symbolic reasoning. Moreover, by continuously extracting potentially correct prefixes, this method has minimal impact on task accuracy and can even enhance it.

%Improving the efficiency of the self-consistency method to achieve lightweight computation is the urgent problem this paper seeks to address.

% As the capabilities of LLMs continue to grow, the range of domains they cover also expands. Researchers, in their quest to explore the limits of LLMs, often conduct experiments in ideal environments with ample computational resources and relaxed conditions. However, due to various limitations such as application scenarios, platforms, and computational resources, large models struggle to demonstrate their full capabilities in practical applications. This underscores the significance of this study on the inference efficiency of LLMs. Achieving more lightweight inference computation methods means a broader application space for models.

\vspace{-2mm}
\section{Background}
\vspace{-2mm}
\subsection{Reasoning with Self-Consistency}
Self-consistency \cite{wang2022self} is a sampling strategy distinct from greedy decoding \cite{vaswani2017attention}, significantly enhancing the reasoning performance of language models. This approach generates multiple reasoning paths and aggregates the final output through majority voting.

Earlier research primarily focused on optimizing reasoning for individual tasks \cite{andor2019giving, ran2019numnet, geva2020injecting, pikekos2021measuring}. More recently, numerous inference strategies based on self-consistency have been proposed, particularly using self-evaluation to calibrate LLMs \cite{zhang2023coder, shinn2023reflexion, madaan2024self, paul2024refiner}. For example, self-evaluation guided by random beam search \cite{xie2024self} leverages additional language models to search for the optimal inference path. Other methods, such as Deductive Beam Search \cite{zhu2024deductive}, optimize inference by emphasizing the relationships between each reasoning step. However, these approaches often require increased computational resources and sometimes involve extensive model calls to improve task accuracy in LLMs. This results in additional time costs, always making them inefficient in practical applications and contrary to our original intention.
\vspace{-1mm}
\subsection{Approaches for Efficient Reasoning}

Recent works have attempted to achieve efficient reasoning by using smaller models or increasing inference speed. Distillation \cite{hinton2015distilling}, as an effective model compression technique, has been employed to create smaller models for reasoning tasks  \cite{fu2023specializing, ho2023large, magister2023teaching, li2023symbolic}. Additionally, various approaches have sought to improve inference speed by altering inference strategies, thereby avoiding modifications to the model architecture and retraining \cite{aggarwal2023let, liescape}. However, these methods often compromise the diversity inherent in self-consistency, which can negatively impact the quality of the generated outputs. The proposed path-consistency is a model-agnostic method applicable to most common models, including compressed models, to enhance inference efficiency while maintaining task accuracy.
\section{Motivation} \label{sec:characterization}
% Figure \ref{fig:self-consistency} illustrates the reasoning process. 
% \vspace{-5mm}
Self-consistency in large language models (LLMs) involves generating multiple branches of reasoning, each potentially leading to different answers. The final answer is determined through aggregation methods, such as majority voting. Unlike greedy decoding, self-consistency avoids the shortcomings of local optimality and reduces the randomness of single-step sampling.
% \begin{figure}[!htp]
% \begin{center}
% \includegraphics[width=0.8\linewidth]{figures/Characterization.pdf}
% \end{center}
% \caption{The proportion of tokens generated by self-consistency on correct or incorrect inference paths. In most tasks, self-consistency wastes over 25\%, or even 50\%, of tokens on incorrect branches, resulting in significant additional costs.}
% \label{fig:characterization}
% \end{figure}
However, the primary drawback of self-consistency is the significant computational redundancy. With $N$ branches, each answer is derived from $N$ similar but independent inference processes, leading to an $N$-fold increase in computational cost compared to greedy decoding for a single problem. To improve self-consistency, the goal is to achieve similar sampling effects while reducing the time cost of redundant computations.

We propose an intuitive hypothesis: for example, a particular mathematical problem might have five different reasoning paths, $p_1$ to $p_5$. If the model frequently errs on paths $p_1$ to $p_3$ while $p_4$ and $p_5$ are relatively simpler, then full self-consistency wastes significant computational resources on the problematic paths. As shown in \cref{fig:characterization}, statistics on the number of tokens generated during self-consistency across various datasets reveal that over 25\% and sometimes even 50\% of tokens are wasted on incorrect branches. By sampling multiple times only on $p_4$ and $p_5$, we could enhance resource utilization and improve output accuracy. Furthermore, storing limited information from paths $p_4$ and $p_5$ to guide subsequent branch generation could significantly accelerate inference speed and efficiency.

Additionally, self-consistency involves extensive redundant processes without yielding intermediate results, with the final answer only emerging at the end. If useful information could be identified early in the generation process to guide subsequent branching, outcomes might improve. Intuitively, when tackling complex problems, using simple criteria to preliminarily assess the quality or correctness of the current generation during intermediate stages can enhance the effectiveness of subsequent steps. Our method aims to reduce the time wasted on incorrect branches while increasing the efficiency of generating correct inference paths.

\begin{figure}[t]
\centering
\includegraphics[width=0.8\linewidth]{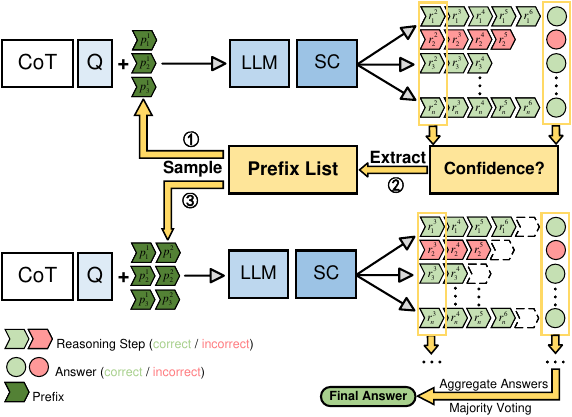}
\caption{An ``extract-and-sample'' inference process of the proposed path-consistency. It seeks the ``optimal path'' in the form of the ``prefix'', thereby progressively reducing the number of generated tokens and significantly shortening inference latency.}
\label{fig:path-consistency}
\vspace{-5mm}
\end{figure}

\section{Methodology}
\vspace{-2mm}
\subsection{Path-Consistency}\label{sec:path-consistency}
\vspace{-2mm}
% \begin{wrapfigure}{r}{0.5\textwidth}
% \centering
% \includegraphics[trim=0cm 0cm 0cm 0cm, clip,height=5cm]{figures/path-consistency-detail-v2.pdf}
% \caption{An ``extract-and-sample'' inference process of the proposed path-consistency. It seeks the ``optimal path'' in the form of the ``prefix''. Thereby progressively reducing the number of generated tokens and significantly shortening inference latency.}
% \label{fig:path-consistency}
% \end{wrapfigure}

Based on the internal mechanisms of self-consistency, we propose an automated dynamic reasoning approach \textit{path-consistency} that continually seeks the ``optimal path'' in the form of ``prefix''. This will progressively reduce the number of generated tokens and significantly shorten inference latency. The methodology is shown in \cref{fig:path-consistency} and \cref{tab:path-consistency}, and can be described as the following ``extract-and-sample'' process:

\begin{itemize}[noitemsep, topsep=0pt, partopsep=0pt]
\item Determine the maximum number of branches and the highest prefix level for the specific task, and use these to divide all branches into multiple windows of equal length. Begin by generating a small number of branches for the first window. 
\item Then, use a confidence assessment metric\footnote{The proposed path-consistency allows a flexible choice of confidence assessment metrics (i.e., stopping criteria), based on intended objective and requirements. The confidence assessment metric used in the evaluation is the beta confidence criteria \cite{aggarwal2023let}.} to assess the confidence of the answers generated in the current window. If the confidence exceeds the set threshold, extract shorter prefixes from these optimal paths (e.g., the first step of the current optimal reasoning paths) to guide subsequent generation; if the confidence is low, continue generating branches for the subsequent windows at the current prefix level to ensure a safer prefix selection.
\item Randomly sample from the extracted prefixes as part of the prompt to guide subsequent generation. After the branches for the next window are generated, continue using the confidence assessment metric to find the optimal inference path and extract the prefix for the next level (e.g., the first two steps of the current optimal reasoning paths).
\item Repeat the above steps iteratively, extending the prefix length until the optimal reasoning path is identified and aggregated to produce the final answer.
\end{itemize}

\Cref{tab:path-consistency} illustrates the reasoning process of path-consistency with an example. We have taken a reasoning path from different stages of the branching process as samples in \cref{tab:path-consistency}. In the initial stage, the model generates the full reasoning paths and answers normally. When a particular answer appears frequently and has a higher confidence level compared to other answers within the current branch, the first step of the corresponding reasoning path is extracted and used as the input prompt for the next stage. As shown in \cref{tab:path-consistency}, as the number of branches increases, the prefix length grows while the newly generated portion of the path subsequently shortens.

\begin{table}[thp]
    % \vspace{-0.2cm}
    \setlength{\abovecaptionskip}{0.1cm}
    \caption{The generated parts are indicated in \color{blue}blue \color{black}in the table, with ``Level-$X$ prefix'' denoting the action of extracting the prefix for the $X^{th}$ time. After three prefix extractions, the model only needs to generate the final sentence of the conclusion based on the prefix.}
    \label{tab:path-consistency}
    \centering
    \small
    % \begin{tabular}{@{}l@{}}
    \begin{tabularx}{0.95\columnwidth}{@{}X@{}}
        \toprule
        
        \textbf{CoT-prompting:} \\
        Q: A robe takes 2 bolts of blue fiber and half that much white fiber. How many bolts in total does it take? \\
        A: \color{blue}{To make a robe, we need 2 bolts of blue fiber and half that much white fiber. We can write this as 2 + 1/2 * 2. Now 1/2 * 2 = 1, so the equation becomes 2 + 1 = 3. The answer is 3.} \\\\
        
        \textbf{Level-I prefix:} \\
        Q: A robe takes 2 bolts of blue fiber and half that much white fiber. How many bolts in total does it take? \\ 
        A: To make a robe, we need 2 bolts of blue fiber and half that much white fiber. \color{blue}{2 * 1/2 = 1 bolt of white fiber. 2 + 1 = 3 bolts in total. The answer is 3.} \\\\
        
        \textbf{Level-II prefix:} \\
        Q: A robe takes 2 bolts of blue fiber and half that much white fiber. How many bolts in total does it take? \\ 
        A: To make a robe, we need 2 bolts of blue fiber and half that much white fiber. 2 * 1/2 = 1 bolt of white fiber. \color{blue}{1 + 2 = 3 bolts in total. The answer is 3.} \\\\        
        \textbf{Level-III prefix:} \\
        Q: A robe takes 2 bolts of blue fiber and half that much white fiber. How many bolts in total does it take? \\ 
        A: To make a robe, we need 2 bolts of blue fiber and half that much white fiber. 2 * 1/2 = 1 bolt of white fiber. 1 + 2 = 3 bolts in total. \color{blue}{The answer is 3.} \\ \bottomrule
    \end{tabularx}
    % \end{tabular}
    % \vspace{-0.3cm}
\end{table}

When solving complex problems, LLM takes a question, represented as $q$, and generates a distribution of answers, denoted as $P(a\mid q)$, after producing a reasoning path. During inference, the model generates a multi-step reasoning path $ R = \left[ r^1, r^2, \cdots, r^T \right ] = r^{1:T}$  under the guidance of CoT. This process can also be expressed as 
% \vspace{-0.5cm}
\begin{equation}
    P(a\mid q)=\mathbb{E}_{R\sim P(R\mid q)}P(a\mid q,R).
\end{equation}
% \vspace{-0.5cm}

In basic self-consistency, the generation process is mechanically repeated numerous times without variation, whereas path-consistency attempts to gather advantageous intermediate information from the early stages of the generation process for subsequent generations. Path-consistency gradually identifies sufficiently confident partial reasoning steps $R_\text{prefix} = [r^1, r^2, \cdots, r^t]$ and incorporates them as part of the input, thereby better guiding the subsequent reasoning steps and the generation of the answer $a$. Thus, this process can be expressed as
% % \vspace{-0.5cm}
\begin{equation}
\small
    P(a\mid q, R_\text{prefix})=\mathbb{E}_{R\sim P(R\mid q, R_\text{prefix})}P(a\mid q, R).
\end{equation}
% \vspace{-0.5cm}

Combining the above expressions, we obtain the following formula via the law of total probability:
% \vspace{-0.5cm}
\begin{equation}
\small
    P(a\mid q)=\sum_{R_{\text{prefix}}}P(R_{\text{prefix}}\mid q)P(a\mid q, R_{\text{prefix}}).
\end{equation}
% \vspace{-0.5cm}
Path-consistency utilizes basic majority voting or other lightweight confidence metrics to model the distribution $P(R_{\text{prefix}}\mid q)$, facilitating the transformation from $P(a\mid q)$ to $P(a\mid q, R_\text{prefix})$. This approach enhances efficiency while maintaining the model's performance across various tasks.

\subsection{Problems with ``Truth Is in Hands of a Few''}

In the basic self-consistency method, when generating answers for particularly challenging questions, the problems with ``Truth Is in the Hands of a Few'' may still occur despite the repeated generation of multiple branches. This means that the correct answer may not be the most frequently occurring one, leading to an incorrect final answer in majority voting. In the proposed path-consistency method, during the continuous exploration for the optimal path, if the problems with ``Truth Is in the Hands of a Few'' are encountered at a specific prefix selection, there's a concern that this undesirable phenomenon may be exacerbated in subsequent branches. We will use the following to demonstrate that the proposed approach does not worsen this problem. 

To present the analysis, we make the following assumptions: the probability of generating the correct answer is $p_0$, and apart from the correct answer, the model generates only one unique incorrect answer. The total number of branches is set to $N$, and prefix selection based on majority voting is performed only once at $\frac{N}{2}$. If the correct answer is the majority at the time of prefix selection, the correct prefix will be selected to guide subsequent generation, increasing the probability of generating the correct answer in the remaining branches to $p_1 = p_0 + \Delta p$; Conversely, if the incorrect answer is the majority at this point, the probability of generating the correct answer in subsequent branches decreases to $p_2 = p_0 - \Delta p$. Using the binomial distribution formula, the probability of the correct answer being in the majority during the vote is given by:
\begin{equation}
    \small
    P_{\text{vote}}=\sum_{k=\lceil\frac{N}{4}\rceil}^{\frac{N}{2}}\binom{\frac{N}{2}}{k}p_0^k(1-p_0)^{\frac{N}{2}-k}.
\end{equation}

After prefix selection, the probability of obtaining the correct answer in the subsequent $\frac{N}{2}$ branches increases or decreases to
{
% scriptsize
\begin{scriptsize}
\begin{equation}
    \begin{aligned}\label{eq:p_inc}
        P_{\text{inc}} =&\sum_{k=\lceil\frac{N}{4}\rceil}^{\frac{N}{2}}\binom{\frac{N}{2}}{k}p_1^k(1-p_1)^{\frac{N}{2}-k} \\
          = &\sum_{k=\lceil\frac{N}{4}\rceil}^{\frac{N}{2}}\binom{\frac{N}{2}}{k}(p_0 + \Delta p)^k(1-p_0-\Delta p)^{\frac{N}{2}-k} \\
         = &\sum_{k=\lceil\frac{N}{4}\rceil}^{\frac{N}{2}}\binom{\frac{N}{2}}{k}p_0^k\left(1+\frac{\Delta p}{p_0}\right)^k(1-p_0)^{\frac{N}{2}-k}\left(1-\frac{\Delta p}{1-p_0}\right)^{\frac{N}{2} - k} \\
        \approx &  \sum_{k=\lceil\frac{N}{4}\rceil}^{\frac{N}{2}}\binom{\frac{N}{2}}{k}p_0^k(1-p_0)^{\frac{N}{2}-k}\left(1+\frac{k}{p_0}\Delta p\right)\left(1-\frac{\frac{N}{2}-k}{1-p_0}\Delta p\right) \\
        \approx & \sum_{k=\lceil\frac{N}{4}\rceil}^{\frac{N}{2}}\binom{\frac{N}{2}}{k}p_0^k(1-p_0)^{\frac{N}{2}-k}\left(1+\frac{k-\frac{N}{2}p_0}{p_0(1-p_0)}\Delta p\right),
\end{aligned}
\end{equation}
\end{scriptsize}
}
and
\begin{scriptsize}
\begin{equation}\label{eq:p_dec}
    % \small
    \begin{aligned}
        P_{\text{dec}}= & \sum_{k=\lceil\frac{N}{4}\rceil}^{\frac{N}{2}}\binom{\frac{N}{2}}{k}p_2^k(1-p_2)^{\frac{N}{2}-k} \\
        = & \sum_{k=\lceil\frac{N}{4}\rceil}^{\frac{N}{2}}\binom{\frac{N}{2}}{k}(p_0 - \Delta p)^k(1-p_0 + \Delta p)^{\frac{N}{2}-k} \\
        \approx & \sum_{k=\lceil\frac{N}{4}\rceil}^{\frac{N}{2}}\binom{\frac{N}{2}}{k}p_0^k(1-p_0)^{\frac{N}{2}-k}\left(1-\frac{k-\frac{N}{2}p_0}{p_0(1-p_0)}\Delta p\right).
    \end{aligned}
\end{equation}
\end{scriptsize}

The two equations above use the first-order Taylor expansion, neglecting higher-order terms. The relationship between the two values is
\begin{equation} \label{eq:prove}
    P_{\text{inc}}+P_{\text{dec}}=2\cdot P_{\text{vote}}.
\end{equation}
Meanwhile, the final probability of obtaining the correct answer after prefix selection is
\begin{equation}
    P_\text{{correct}}^{\prime} = P_{\text{vote}}\cdot P_{\text{inc}} + (1 - P_{\text{vote}})\cdot P_{\text{dec}}.
\end{equation}

If no prefix selection is performed, the probability of generating the correct answer in subsequent steps remains unchanged and is equal to the probability at $\frac{N}{2}$ with majority voting: $P_\text{{correct}} =P_{\text{vote}}$.
% \begin{equation}
%     P_\text{{correct}} =P_{\text{vote}},
% \end{equation}
To ensure that accuracy is not adversely affected, we require $P_\text{{correct}}^{\prime}\geq P_{\text{correct}}$, which can be simplified to:
% \begin{equation}
%     P_\text{{correct}}^{\prime}\geq P_{\text{correct}},
% \end{equation}
% It can be simplified to:
\begin{equation}
    P_{\text{vote}}\geq \frac{P_{\text{vote}}-P_{\text{dec}}}{P_{\text{inc}}-P_{\text{dec}}}.
\end{equation}
By combining \cref{eq:prove}, we can obtain $P_{\text{vote}}\geq 0.5$. This proves that if the model's initial performance ensures $P_{\text{vote}}\geq 0.5$, path-consistency will be sufficiently reliable and will not harm accuracy.
% From this formula, we observe that the greater the difference between $P_{\text{inc}}$ and $P_{\text{dec}}$, the more reliable this method becomes, and the smaller the impact on accuracy.

During prefix selection, employing confidence-based criteria can make this process more reliable. For instance, using the beta confidence criteria \cite{aggarwal2023let}
\begin{equation}\label{eq:confidence}
    \int_0^{0.5}p^{\frac{N}{2} - v_m}\cdot(1-p)^{v_m}dp,
\end{equation}
where $v_m$ represents the number of major elements. A prefix is considered sufficiently reliable for selection only if this value exceeds a predefined confidence threshold $C_{threshold}$. 

% If we roughly assume $N = 20$, $p_0=0.4$, $\Delta p = 0.01$, the final accuracy without prefix selection is $P_\text{{correct}} = 16.62\%$, and the final accuracy after prefix selection by majority voting is $P_\text{{voting}} = 15.15\%$, by Beta Stopping Criteria is $P_{C = 0.8} = 16.10\%$. Actually, $P_\text{{voting}} = P_{C = 0}$. Under the same conditions, if $p_0 = 0.6 (\ge 0.5)$, then both $P_{C = 0}$ and $P_{C = 0.8}$ will be greater than $P_\text{{correct}}$.

While this calculation is a rough estimate and the actual scenario is likely more complex, it provides insight into the effect of path-consistency on problems where ``Truth Is in the Hands of a Few''. Essentially, path-consistency tends to enhance the accuracy of self-consistency: when self-consistency performs well for a particular input, path-consistency may perform even better; when self-consistency performs poorly, the accuracy of path-consistency might decrease, but not significantly. Moreover, prefix selection guided by a confidence threshold provides a safer alternative to direct majority-vote selection.

\section{Evaluation} % Experiments

\subsection{Experimental Setting}

% \subsubsection{Benchmarks.}
\textbf{Benchmarks:}
We evaluated the performance and efficiency improvement of path-consistency across the following types of tasks: (1) Arithmetic Reasoning, including GSM8K \cite{cobbe2021training}, SVAMP \cite{patel2021nlp}, ASDiv \cite{miao2020diverse}, and MultiArith \cite{roy2015solving}; (2) Commonsense Reasoning, including StrategyQA \cite{geva2021did}, Ruin Names, and Salient Translation; (3) Symbolic Reasoning, including Boolean Expressions, Tracking Shuffled Objects, as well as Logical Deduction \cite{srivastava2023beyond}.

% \subsubsection{Model.} 
\textbf{Model:} We used the open-source model Llama3-8B \cite{dubey2024llama} as our backbone model. During input, we employed prompts similar to those used in CoT \cite{wei2022chain}.

% 推理时我们使用
% \subsubsection{Hyperparameters.} 
\textbf{Hyperparameters:} For LLM inference with self-consistency and path-consistency, we employed nucleus sampling with a temperature of 0.6 and top-p of 0.9, generating 20 paths for each example. According to \citet{wang2022self}, 20 sampled reasoning paths almost achieve convergence in task accuracy, with only marginal efficiency gains observed when increasing to 40 paths. Therefore, 20 paths are more suitable for evaluating the optimization capability of path-consistency in terms of efficiency. Meanwhile, we used the beta confidence criteria in \cref{eq:confidence} for confidence calculation in path-consistency \cite{aggarwal2023let}, with threshold values ranging from 0.5 to 1.0. 
%A confidence threshold ($C_{threshold}$) set to 1.0 serves as the baseline for self-consistency, while a threshold set to 0 implies a prefix selection strategy without confidence protection. Additionally, we also conduct experiments with thresholds of 0.7, 0.8 and 0.9.

% \subsubsection{Metrics.} 
\textbf{Metrics:} We employed multiple metrics to comprehensively compare the performance of path-consistency against baselines, including reasoning accuracy, inference latency, and token consumption during reasoning. Since inference latency is highly dependent on the operating environment and hardware configuration, considering the absolute value of inference latency is not meaningful. Instead, we calculated speedup under identical conditions as a standardized evaluation metric. All experiments were conducted on a single NVIDIA GeForce RTX 3090 GPU.
% Experimental results demonstrate that path-consistency achieves task reasoning accuracy comparable to baseline methods while significantly reducing both time and computational costs.

% \subsection{Performance Analysis}

% \vspace{-5mm}
\subsection{Evaluation of Path-Consistency}
\subsubsection{A Case Study on GSM8K} % Efficient Inference Process. zoom-in
\vspace{-5mm}
\begin{figure*}[htbp]
    \begin{minipage}{0.68\textwidth}
    \setlength{\abovecaptionskip}{0.1cm}
    \captionsetup{font=small}
    \captionof{table}{Comparison of inference latency speedup and average token consumption reduction under different prefix levels, demonstrating the effect of path-consistency on GSM8K.}
    \resizebox{\columnwidth}{!}{
        \begin{tabular}{llccccccccc}
        \toprule
                                            &                                      & \multicolumn{4}{c}{Speedup (\%)}           & \multicolumn{1}{l}{}         & \multicolumn{4}{c}{Decrease (\%)}          \\ \cmidrule(lr){3-6} \cmidrule(l){8-11} 
        Method                              & \multicolumn{1}{c}{Accuarcy (\%)}    & Total & Level-1 & Level-2 & Level-3        & \multicolumn{1}{l}{ Tokens (\#)} & Total & Level-1 & Level-2 & Level-3       \\ \midrule
        SC                                  & 64.1                                 &       &         &         &                & 96.60                        &       &         &         &                \\
        PC \small{($C$=0)}   & 66.6 \small{(+2.5)} & +28.9 & +18.8   & +42.6   & \textbf{+78.9} & 73.37                        & -24.0 & -16.9   & -32.7   & \textbf{-47.4} \\
        PC \small{($C$=0.7)} & 67.1 \small{(+3.0)} & +18.0 & +10.0   & +25.3   & \textbf{+46.5} & 81.09                        & -16.1 & -9.4    & -21.3   & \textbf{-33.8} \\
        PC \small{($C$=0.8)} & 67.8 \small{(+3.7)} & +17.4 & +9.8    & +25.0   & \textbf{+43.9} & 80.91                        & -16.2 & -9.8    & -21.8   & \textbf{-33.3} \\
        PC \small{($C$=0.9)} & 66.6 \small{(+2.5)} & +10.2 & +4.2    & +13.3   & \textbf{+27.0} & 87.01                        & -9.9  & -4.2    & -12.7   & \textbf{-23.0} \\ \bottomrule
        \end{tabular}
    }
    \label{tab:speedup}
    \end{minipage}
    \begin{minipage}{0.3\textwidth}
        % \subcaption{speedup}
        \centering
        \setlength{\abovecaptionskip}{0.1cm}
        \includegraphics [width =\textwidth]{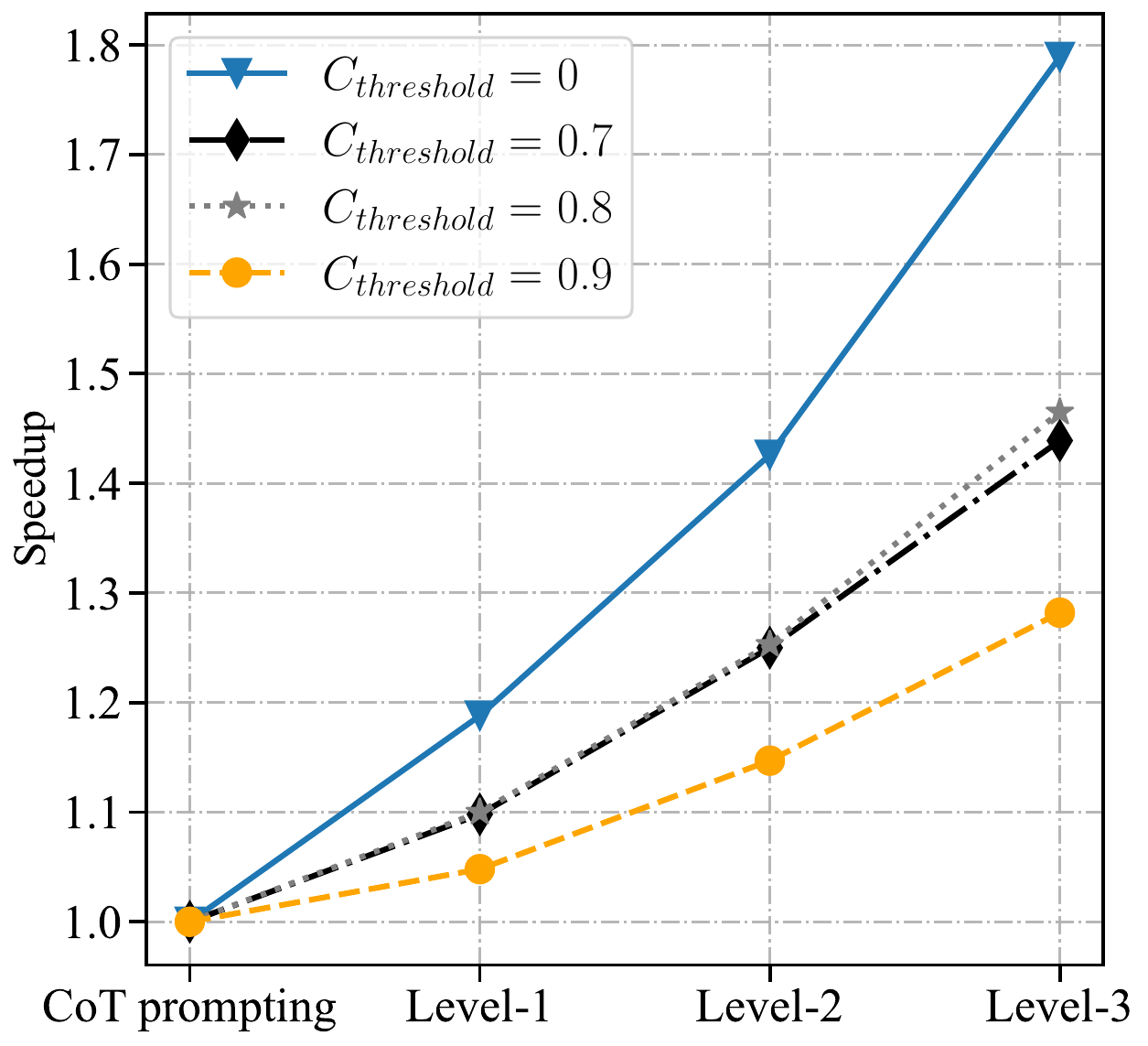}
        \captionsetup{font=small}
        \captionof{figure}{Speedup of inference.}
        \label{fig:speedup}
    \end{minipage}
    % \begin{minipage}{0.22\textwidth}
    %     \centering
    %     \setlength{\abovecaptionskip}{0.1cm}
    %     \includegraphics [width =\textwidth]{figures/token.pdf}
    %     \captionsetup{font=small}
    %     \captionof{figure}{Token consumption.}
    %     \label{fig:token}
    % \end{minipage}
    % \vspace{-0.5cm}
\end{figure*}
\vspace{-4mm}

\Cref{tab:speedup} presents the performance and efficiency of path-consistency on the GSM8K dataset across different confidence thresholds and prefix levels. The experiments compare path-consistency with the basic self-consistency method in terms of task accuracy, inference latency, and token consumption. The results show that, across all confidence criteria, path-consistency not only enhances task accuracy but also provides a maximum acceleration of 28.9\%. Observing the acceleration process of path-consistency, it is evident that as the prefix level increases, the speed improvement becomes more significant, reaching up to 78.9\% at level-3 while maintaining at least 27.0\%, as shown in \cref{fig:speedup}. 
% It can also lead to a 23.0\% to 47.4\% decrease in the number of generated tokens, as shown in \cref{fig:token}.

\subsubsection{Performance on Different Tasks}

\begin{table*}[b]
\centering
\setlength{\abovecaptionskip}{0.1cm}
\caption{The evaluation results of path-consistency across various tasks, including (i) Arithmetic reasoning: GSM8K, SVAMP, ASDiv, MultiArith (MA); (ii) Commonsense reasoning: StrategyQA (SQA), RuinNames (RN), SalientTranslation (ST); (iii) Symbolic reasoning: BooleanExpression (BE), TrackingShuffledObjects (TSO), LogicalDeduction (LD). $C$=0 indicates that no confidence threshold is applied; the prefix corresponding to the majority element is extracted directly.}
\label{tab:result}
\resizebox{\textwidth}{!}{
\begin{tabular}{@{}cccccccccccccc@{}}
\toprule
Dataset&Metrics (\%)&SC&PC \small{($C$=0)}&PC \small{($C$=0.7)}&PC \small{($C$=0.8)}&PC \small{($C$=0.9)}&Dataset&Metrics (\%)&SC&PC \small{($C$=0)}&PC \small{($C$=0.7)}&PC \small{($C$=0.8)}&PC \small{($C$=0.9)} \\
      \midrule
      & Acc.     & 64.1 & 66.6 \small{(+2.6)} & 67.1 \small{(+3.1)}  & \textbf{67.8 \small{(+3.8)}} & 66.6 \small{(+2.6)} &    & Acc.     & 76.0 & 71.2 \small{(-4.8)}  & \textbf{74.4 \small{(-1.6)}} & 72 \small{(-4.0)} & 71.6 \small{(-4.4)} \\
GSM8K & Speedup  & -    & +28.9               & +18.0                & +17.4                        & +10.2               & RN & Speedup  & -    & +10.4                & +7.8                         & +7.4              & +4.7        \\
      & Decrease & -    & -24.0               & -16.1                & -16.2                        & -9.9                &    & Decrease & -    & -9.8                 & -7.2                         & -6.6              & -4.7        \\ 
      \cmidrule(lr){2-7} \cmidrule(l){9-14} 
      & Acc.     & 79.3 & 78.8 \small{(-0.5)} & 78.5 \small{(-0.8)}  & 79.2 \small{(-0.1)} & \textbf{79.9 \small{(+0.6)}} &    & Acc.     & 54.8 & \textbf{58.4 \small{(+3.6)}} & 57.2 \small{(+2.4)} & 56.4 \small{(+1.6)} & 55.2 \small{(+0.4)} \\
SVAMP & Speedup  & -    & +48.3               & +36.6                & +35.7               & +24.8                        & ST & Speedup  & -    & +30.5                        & +19.6               & +17.3               & +10.2       \\
      & Decrease & -    & -36.6               & -29.4                & -29.7               & -22.3                        &    & Decrease & -    & -25.8                        & -18.3               & -16.4               & -10.6       \\
      \cmidrule(l){2-14} 
      & Acc.     & 81.0 & 80.1 \small{(-0.9)} & \textbf{81.2 \small{(+0.2)}} & 80 \small{(-1.0)}  & 80.6 \small{(-0.4)}   &    & Acc.     & 88.4 & \textbf{90 \small{(+1.6)}}   & 89.6 \small{(+1.2)} & 88.8\small{(+0.4)}  & 89.6 \small{(+1.2)} \\
ASDiv & Speedup  & -    & +42.0               & +32.1                        & +31.1              & +24.9                 & BE & Speedup  & -    & +26.6                        & +24.3               & +25.2               & +22.0       \\
      & Decrease & -    & -32.7               & -27.2                        & -27.1              & -22.6                 &    & Decrease & -    & -21.3                        & -19.8               & -20.3               & -18.3       \\
      \cmidrule(lr){2-7} \cmidrule(l){9-14} 
      & Acc.     & 96.7 & 97.8 \small{(+1.1)} & 97.2 \small{(+0.5)}  & \textbf{98.9 \small{(+2.2)}} & 98.3 \small{(+1.6)} &    & Acc.     & 54.0 & 58 \small{(+4.0)}  & 56 \small{(+2.0)} & \textbf{59.2 \small{(+5.2)}} & 58.4 \small{(+4.4)}  \\
MA    & Speedup  & -    & +42.7               & +40.8                & +40.5                        & +33.8               & TSO& Speedup  & -    & +18.4              & +12.9             & +11.3                        & +6.8        \\
      & Decrease & -    & -34.8               & -33.3                & -33.1                        & -27.2               &    & Decrease & -    & -16.7              & -12.4             & -10.8                        & -6.8        \\
      \cmidrule(r){1-7} \cmidrule(l){9-14} 
      & Acc.     & 71.4 & 70.8 \small{(-0.6)} & 70.6 \small{(-0.8)}  & \textbf{71.3 \small{(-0.1)}} & 71.1 \small{(-0.3)} &    & Acc.     & 78.8 & 76.8 \small{(-2,0)} & \textbf{78.8 \small{(+0.0)}} & 77.2 \small{(-1.6)} & 76 \small{(-2.8)}   \\
SQA   & Speedup  & -    & +34.2               & +29.4                & +24.3                        & +16.4               & LD & Speedup  & -    & +31.6               & +23.1                        & +30.1               & +19.1       \\
      & Decrease & -    & -27.8               & -24.0                & -22.1                        & -16.0               &    & Decrease & -    & -26.1               & -20.5                        & -24.8               & -17.5       \\
      \bottomrule
\end{tabular}
}
% \vspace{-0.5cm}
\end{table*}
\textbf{Arithmetic Reasoning:}
% \subsubsection{Arithmetic Reasoning.}
\Cref{tab:result}, rows 1-4, shows the task performance on arithmetic reasoning datasets, along with three additional datasets. Path-consistency ensures task performance that is almost comparable to or even better than the baseline under various confidence settings. On GSM8K, SVAMP, ASDiv, and MultiArith, accuracy improvements of up to 3.8\%, 0.6\%, 0.2\%, and 2.2\% are observed, respectively. Due to the varying difficulty levels of the datasets for the LLM, the optimal confidence threshold differs for each dataset. For instance, higher or lower confidence thresholds may enhance performance on SVAMP and ASDiv.

The results also highlight the efficiency improvements on arithmetic reasoning datasets. Path-consistency achieves acceleration rates of up to 28.9\%, 48.3\%, 42.0\%, and 42.7\% on GSM8K, SVAMP, ASDiv, and MultiArith, respectively, along with a corresponding decrease in token consumption. Comparing these results reveals that for a specific dataset, there exists a proper confidence threshold at which path-consistency simultaneously enhances both task performance and efficiency.

\textbf{Commonsense Reasoning:}
% \subsubsection{Commonsense Reasoning.}
\Cref{tab:result}, rows 5-7, reports the performance of path-consistency on commonsense reasoning datasets. In terms of task accuracy, path-consistency slightly underperforms compared to the baseline on the StrategyQA and Ruins Names tasks. However, this decline can be mitigated by properly adjusting the confidence threshold. On the Salient Translation task, path-consistency still maintains better task performance compared to the baseline. Regarding acceleration, path-consistency provides up to 34.2\%, 10.4\%, and 30.5\% speed improvements on StrategyQA, Ruins Names, and Salient Translation datasets, respectively.

According to the results, it can be found that the performance of path-consistency on the Ruins Names task is less pronounced compared to other datasets. This is because the Ruins Names task is more difficult, with longer input content and longer required prompts, making the benefits of the prefixing behavior less noticeable.

\begin{wrapfigure}{r}{0.5\textwidth}
\centering
\vspace{-6mm}
\includegraphics[trim=0cm 0cm 0cm 0cm, clip,height=45mm]{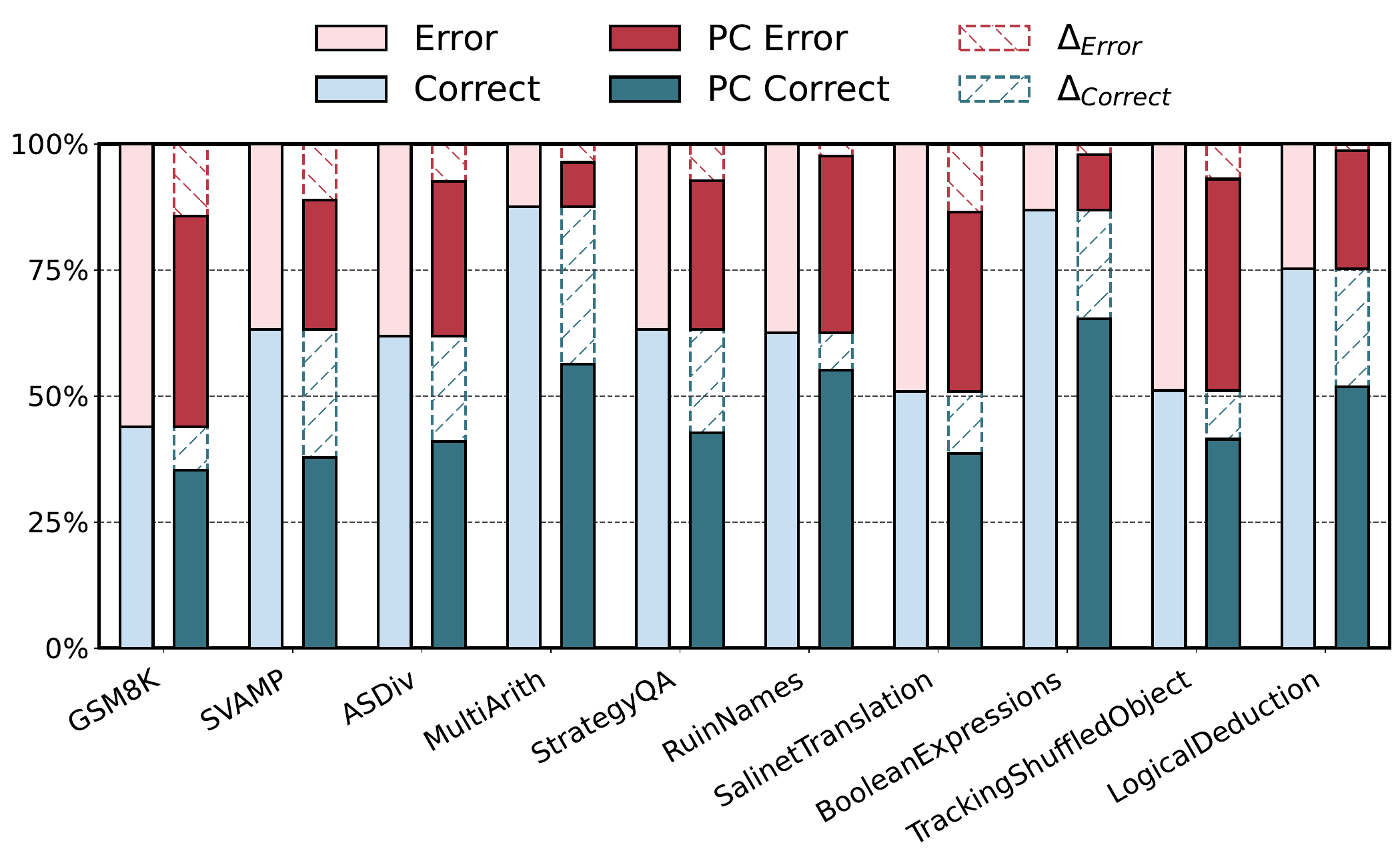}
\vspace{-8mm}
\caption{The change in the proportion of tokens generated by path-consistency on correct or incorrect paths.}
\label{fig:Error and Redundant}
\vspace{-6mm}
\end{wrapfigure}

\textbf{Symbolic Reasoning:}
% \subsubsection{Symbolic Reasoning.}
\Cref{tab:result}, rows 8-10, compares the performance of different methods on the symbolic reasoning datasets. Path-consistency performs exceptionally well in terms of task accuracy, especially in the Tracking Shuffled Objects task, where it achieves the highest improvement of 5.2\% among all datasets. Additionally, compared to the baseline, it delivers approximately a 20\% speedup in inference latency across tasks such as Boolean Expressions, Tracking Shuffled Objects, and Logical Deduction.

In general, lower confidence thresholds indicate a more aggressive prefix selection strategy, often resulting in more significant efficiency improvements. However, in the Logical Deduction task, the efficiency improvement is actually higher at a confidence threshold of 0.8 compared to 0.7. This is because a more aggressive prefix selection strategy is more likely to choose suboptimal prefixes, which makes the subsequent branching generation less effective. Specifically, we found that with a confidence threshold of 0.7, the efficiency improvement at prefix level-2 is more significant than at prefix level-3, as prefix level-3 generates many erroneous paths, leading to this phenomenon.

\subsubsection{Result Analysis}

\begin{wraptable}{r}{0.5\textwidth}
\vspace{-5mm}
  \caption{Impacts of maximum path levels of path-consistency on GSM8K.}
  \label{tab:level}
\vspace{-1mm}
\begin{scriptsize}
\setlength{\tabcolsep}{4pt}
    \begin{tabular}{@{}ccccc@{}}
    \toprule
    \multicolumn{1}{l}{Method} & \multicolumn{1}{l}{Level} & \multicolumn{1}{l}{Accuracy (\%)} & \multicolumn{1}{l}{Sqeedup (\%)} & \multicolumn{1}{l}{Decrease (\%)} \\ \midrule
    PC ($C$=0.7)                       & 3                              & 67.1                        & +18.0                              & -16.1                            \\
                                             & 4                              & 66.3                        & +20.5                              & -20.0                            \\ \midrule
    PC ($C$=0.8)                       & 3                              & 67.8                        & +17.4                              & -16.2                            \\
                                             & 4                              & 67.8                        & +18.0                              & -17.8                            \\ \bottomrule
    \end{tabular}

\end{scriptsize}
\end{wraptable}%

\textbf{Reducing Error and Redundancy:}
In \cref{sec:characterization}, we characterized the challenges of self-consistency. On one hand, it wastes computational resources on incorrect branches. On the other hand, it repeats the same computations without obtaining useful intermediate information. \Cref{fig:Error and Redundant} shows the changes in the proportion of tokens generated by path-consistency on correct or incorrect reasoning paths. The results show a decrease in the number of tokens wasted on incorrect branches. Additionally, for correct branches, there is a significant reduction in redundant tokens, which greatly improves efficiency while maintaining accuracy.

\textbf{Hyperparameters Analysis:}
We explored the impact of prefix extraction frequency on path-consistency. With the setting of 20 branches, if the highest prefix level is set to level-3, the prefix is extracted every 5 branches; if the highest prefix level is level-4, the prefix is extracted every 4 branches. \Cref{tab:level} indicates that maintaining the highest prefix level at level-3 preserves higher task accuracy. Increasing the highest level enables a more aggressive strategy, leading to a more noticeable acceleration.

\subsection{Comparison with Similar Work} 

We compared path-consistency (PC) to the following methods: (1) \textit{Self-consistency} (SC) \cite{wang2022self} is the baseline that combines multiple samplings with majority voting. (2) \textit{Adaptive-consistency} (AC) \cite{aggarwal2023let} introduces the concept of confidence for the first time. If a highly confident answer is identified, the sampling process is terminated, and the answer is selected as the final output, preventing any further branch sampling and inference. (3) \textit{Early-stopping self-consistency} (ESC) \cite{liescape} divides branches into equally sized windows. When all the answers generated within a given window are identical, then exit directly.

As shown in the experimental data in \cref{tab:adapt}, the path-consistency achieves task accuracy comparable to other optimization methods while consuming fewer tokens.
The diversity of the reasoning paths is the key to a better performance in self-consistency \cite{wang2022self}. AC and ESC achieve significant acceleration by sacrificing diversity through early exit strategies under large branching. In contrast, path-consistency strives to preserve the diversity of reasoning paths during the sampling process from the prefix set. It makes more efficient use of available information, achieving task accuracy comparable to other methods with fewer branches and tokens.

\begin{table*}[h]
\caption{The performance of various inference techniques on GSM8K, Svamp and StrategyQA.}
\label{tab:adapt}
\resizebox{\textwidth}{!}{
\begin{tabular}{@{}lclclcl@{}}
\toprule
       & \multicolumn{2}{c}{GSM8K}              & \multicolumn{2}{c}{Svamp}              & \multicolumn{2}{c}{StratrgyQA}         \\ \cmidrule(l){2-3}\cmidrule(l){4-5}\cmidrule(l){6-7} 
Method                      & Accuracy (\%) & Tokens (\# per problem) & Accuracy (\%) & Tokens (\# per problem) & Accuracy (\%) & Tokens (\# per problem) \\ \midrule
SC \cite{wang2022self}      & 68.4          & 3,860                   & 80.4          & 2,321                   & 71.7          & 1,805                   \\
AC \cite{aggarwal2023let}   & 67.2          & 2,248 (-41.7\%)         & 79.7          & 935 (-59.7\%)          & 71.2          & 706 (-60.9\%)                   \\
ESC \cite{liescape}         & 67.2          & 2,710 (-29.8\%)         & 79.2          & 1,154 (-50.3\%)         & 70.8          & 782 (-56.7\%)                   \\
PC                          & 67.8          & 1,630 (-57.8\%)         & 79.2          & 816 (-64.8\%)          & 71.3          & 700 (-61.2\%)                   \\ \bottomrule
\end{tabular}
}
\end{table*}

\begin{table*}[h]
\setlength{\abovecaptionskip}{0.1cm}
\caption{Evaluation results of different inference techniques on DeepSeek-V3.}
\label{tab:mistral}
\centering
\resizebox{\textwidth}{!}{
\begin{tabular}{@{}lclclcl@{}}
\toprule
                                           & \multicolumn{2}{c}{GSM8K}              & \multicolumn{2}{c}{Svamp}              & \multicolumn{2}{c}{StratrgyQA}         \\ \cmidrule(l){2-3} \cmidrule(l){4-5} \cmidrule(l){6-7}  
Method                                     & Accuracy (\%) & Tokens (\# per problem) & Accuracy (\%) & Tokens (\# per problem) & Accuracy (\%) & Tokens (\# per problem) \\ \midrule
PC                                         & 91.8          & 786                    & 94.7          & 263                    & 84.5          & 230                    \\
AC \cite{aggarwal2023let} & 91.6          & 828 (+5.34\%)          & 94.7          & 363 (+38.0\%)          & 84.5          & 326 (+41.7\%)          \\
ESC \cite{liescape}       & 91.6          & 868 (+10.4\%)          & 94.7          & 358 (+36.1\%)                       & 84.2              & 315 (+36.9\%)                  \\\bottomrule
\end{tabular} 
}
\end{table*}

\vspace{-3mm}
\subsection{Discussion on Robustness and Scalability}
% \subsubsection{Code generation}
Path-consistency doesn't alter the model generation process. It is a lightweight, model-agnostic approach that requires no additional computation. As demonstrated in \cref{tab:mistral}, it maintains strong robustness even when applied to another open-source model, DeepSeek-V3 \cite{liu2024deepseek}.

We can also observe the scalability of path-consistency with larger model sizes. Due to the inherently strong reasoning capabilities of larger models, the impact of the three optimization methods on accuracy remains relatively stable. However, path-consistency still achieves a substantial reduction in token consumption. This demonstrates that path-consistency exhibits excellent robustness and scalability, consistently providing efficiency improvements across different models and at larger scales. Furthermore, as shown in the Appendix, we also conducted experiments on a much smaller model.

\section{Conclusion}
\vspace{-2mm}
This paper proposes path-consistency, which achieves internal optimization of self-consistency, extracting information from early branches in the form of ``prefixes'', guiding the generation of subsequent branches more efficiently. The effectiveness of path-consistency across a broad range of tasks, including arithmetic, commonsense, and symbolic reasoning, demonstrates its robustness in various application areas. Additionally, path-consistency is a lightweight, model-agnostic method that requires minimal additional computational resources and scales effectively to larger models.

% Path-consistency effectively enhances LLMs' efficiency in performing reasoning tasks in every task. Experiments reveal that using an appropriate confidence threshold can improve efficiency while ensuring the quality of the generated output and task accuracy.

% \subsection{Limitations}

% The effectiveness of path-consistency is closely related to the generation quality of LLMs. When the task is challenging and the model struggles to solve it effectively, path-consistency may be less effective. Additionally, when the model generates longer and more complex reasoning steps, the acceleration effect of path-consistency may be less pronounced. However, as the capabilities of the model improve, the potential of path-consistency can be better utilized.

% \bibliography{iclr2026_conference}
% \bibliographystyle{iclr2026_conference}

\appendix
% \section{Appendix}
% \newpage
\appendix
\section{Appendix}

\subsection{Further Analysis}
\subsubsection{Performance in small model}

As presented in the \cref{tab:slm}, path-consistency substantially enhances the inference efficiency of small models on reasoning tasks. Nevertheless, due to the inherent accuracy limitations of the base models, the performance of path-consistency is inevitably influenced as well.

\begin{table*}[h]
\setlength{\abovecaptionskip}{0.1cm}
\caption{Evaluation results of different inference techniques on Llama-3.2-1B-Instruct.}
\label{tab:slm}
\centering
\resizebox{\textwidth}{!}{
\begin{tabular}{@{}lclclcl@{}}
\toprule
                                           & \multicolumn{2}{c}{GSM8K}              & \multicolumn{2}{c}{Svamp}              & \multicolumn{2}{c}{StratrgyQA}         \\ \cmidrule(l){2-3} \cmidrule(l){4-5} \cmidrule(l){6-7}  
Method                                     & Accuracy (\%) & Tokens (\# per problem) & Accuracy (\%) & Tokens (\# per problem) & Accuracy (\%) & Tokens (\# per problem) \\ \midrule
PC                                         & 47.3          & 1430                    & 62.0          & 711                    & 63.1          & 799                    \\
AC \cite{aggarwal2023let} & 48.1          & 2748          & 61.1          & 800          & 63.9          & 1401          \\
ESC \cite{liescape}       & 47.9          & 2324          & 61.9          & 924                       & 63.7              & 1136                  \\\bottomrule
\end{tabular} 
}
\end{table*}

\subsubsection{Robustness to Confidence Threshold}

As shown in Figure \ref{fig:confidence-threshold}, we examined the relationship between the confidence of the majority element in the tenth branch of the self-consistency baseline and the final answer across all datasets. It was observed that a confidence level around 0.8 serves as a critical threshold for determining the correctness of the final answer. Therefore, selecting confidence levels in the range of [0.7, 0.8, 0.9] is a more appropriate choice.

\subsubsection{Protection of High-Confidence Answers}

\begin{figure}[htbp]
    \begin{minipage}{0.4\textwidth}
        \centering
        \includegraphics[width =\textwidth]{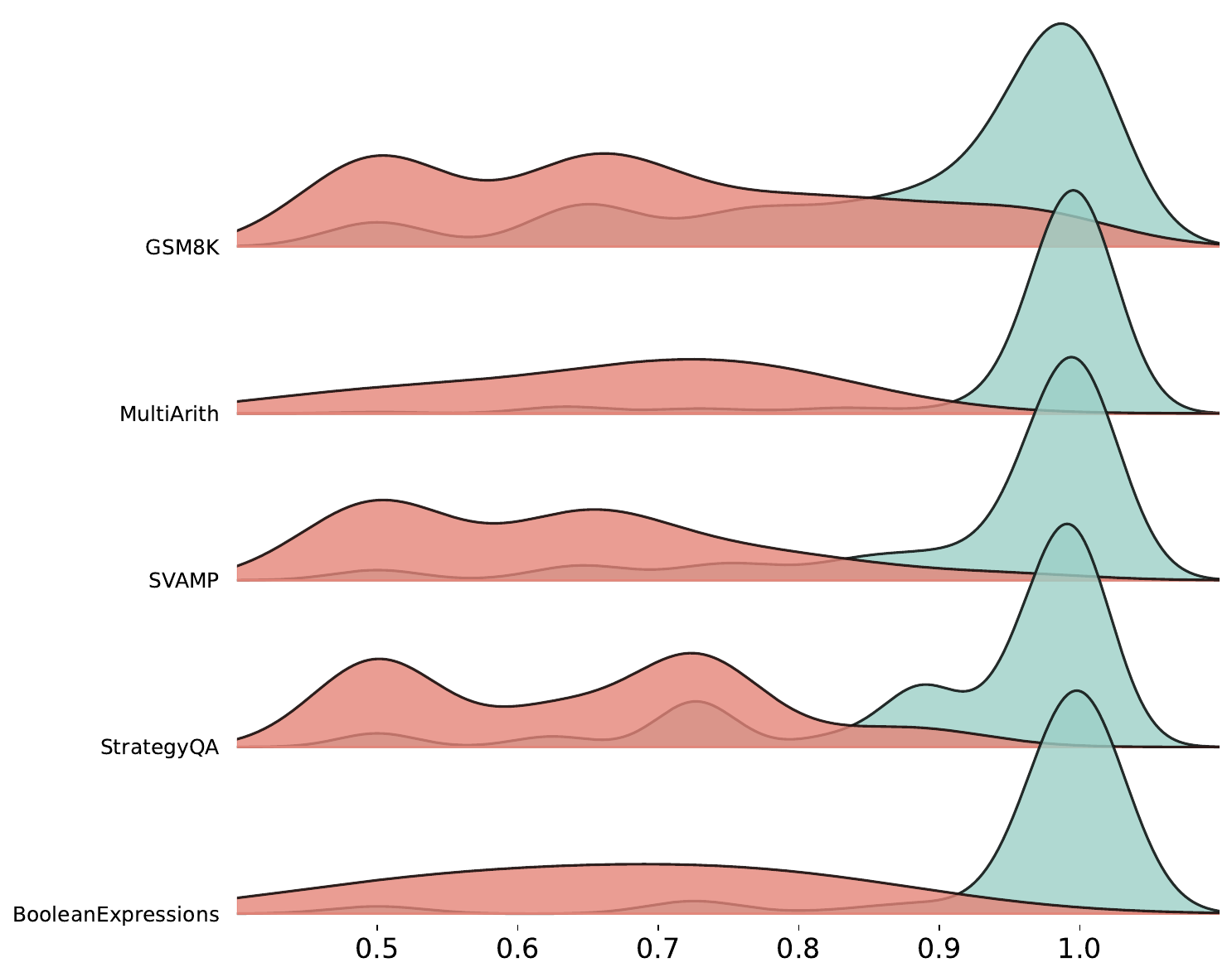}
        \captionof{figure}{The distribution of the majority element's confidence across all samples when self-consistency reaches the 10$^{th}$ branch.}
        \label{fig:confidence-threshold}
    \end{minipage}
    \begin{minipage}{0.55\textwidth}
        \begin{minipage}{0.49\textwidth}
            \centering
            \includegraphics [width =\textwidth]{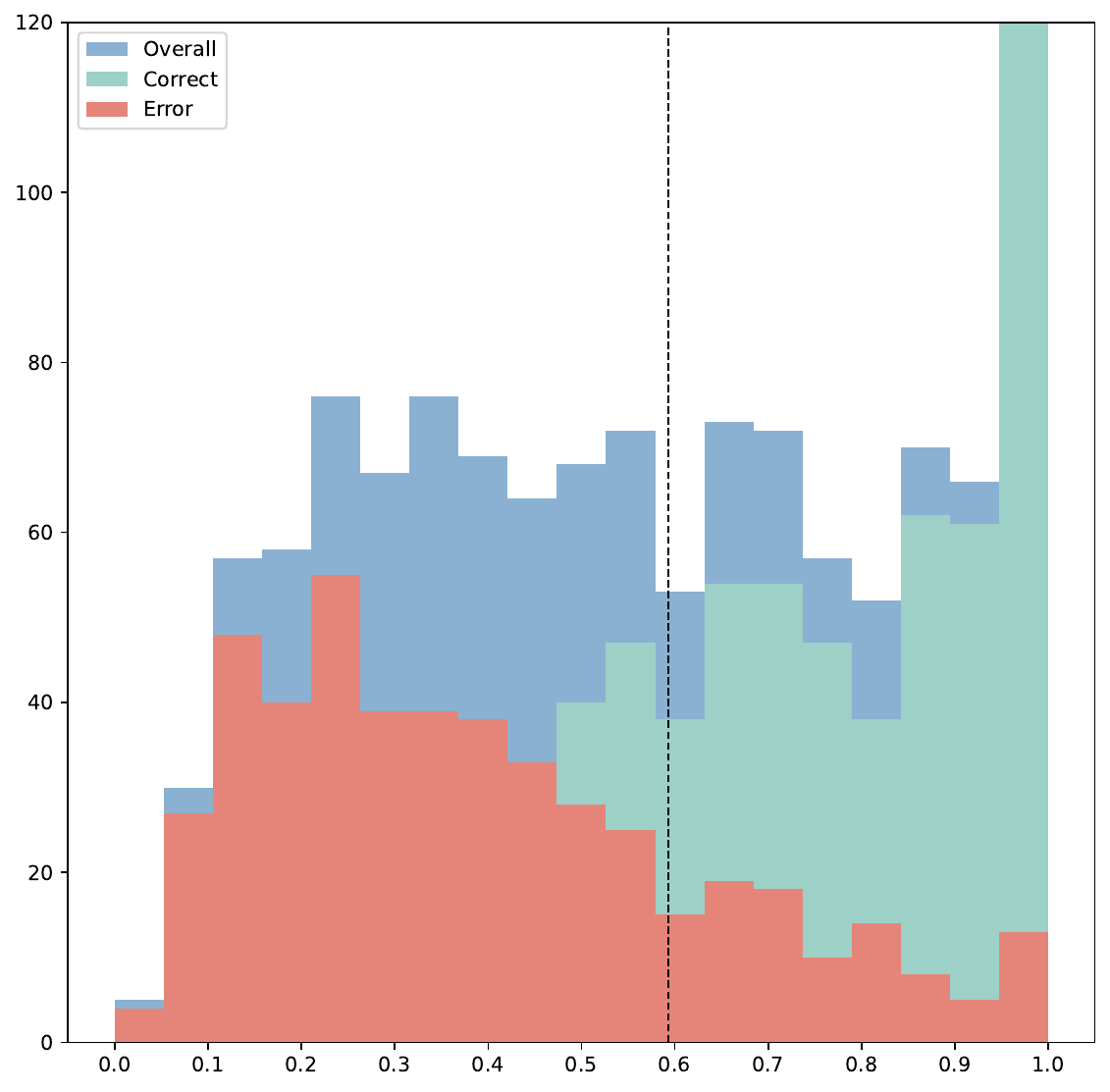}
            \subcaption{Baseline}
            % \captionsetup{font=small}
            % \captionof{figure}{The speedup of inference by path-consistency at different prefix levels.}
            \label{fig:speedup_2}
        \end{minipage}
        \begin{minipage}{0.49\textwidth}
            \centering
            \includegraphics [width =\textwidth]{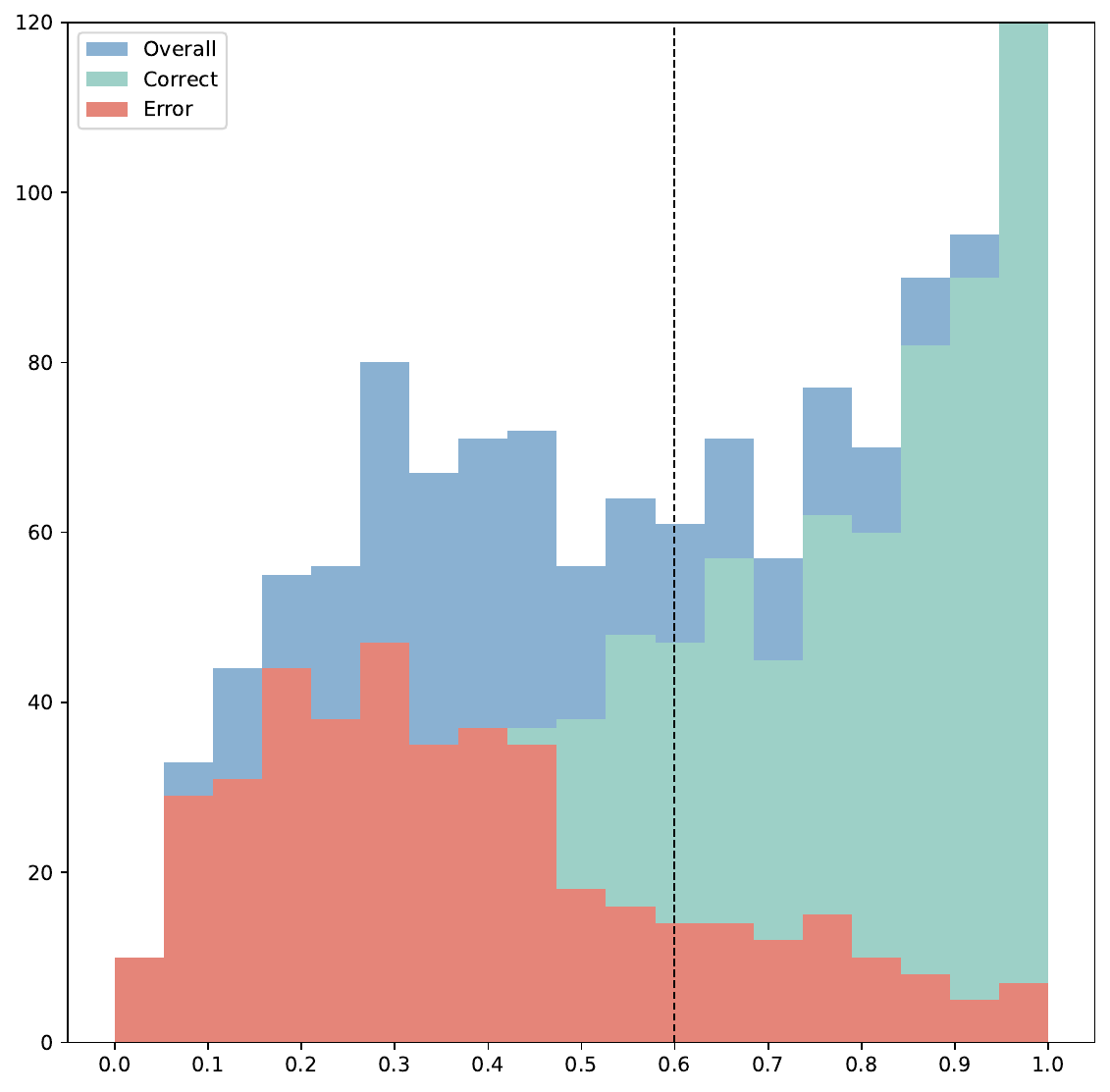}
            \subcaption{$C_{threshold}$ = 0.8}
            % \captionsetup{font=small}
            % \captionof{figure}{The decrease of token number by path-consistency at different prefix levels.}
            \label{fig:token_2}
        \end{minipage}
        \captionof{figure}{The distribution of confidence for the final answers across all samples on GSM8K.}
        \label{fig:distribution}
    \end{minipage}
\end{figure}

As illustrated in the Figure \ref{fig:distribution}, the distribution of confidence for the final answers across all samples is shown. The implementation of path-consistency has led to a significant increase in the number of high-confidence answers, while reducing the number of samples with confidence levels around 0.5. This suggests that path-consistency effectively resolves many ambiguous cases. Consequently, this approach not only avoids compromising task accuracy but may even enhance it.

% \onecolumn
\subsection{Code Analysis} 
Section \ref{sec:path-consistency} introduces the ``extract-and-sample'' process of path-consistency. Following is the code implementation of the key parts:

% Define custom colors (optional)
\definecolor{codebackground}{rgb}{0.95,0.95,0.92}
\definecolor{codeframe}{rgb}{0.8,0.8,0.8}

% Customize the appearance of the code
\lstset{
    backgroundcolor=\color{codebackground}, % Background color
    basicstyle=\ttfamily\footnotesize,      % Font style and size
    frame=single,                           % Adds a frame around the code
    rulecolor=\color{codeframe},            % Frame color
    keywordstyle=\color{blue},              % Keyword color
    commentstyle=\color{gray},              % Comment color
    stringstyle=\color{red},                % String color
    numbers=left,                           % Line numbers on the left
    numberstyle=\tiny\color{gray},          % Line number style
    stepnumber=1,                           % Number every line
    numbersep=5pt,                          % Space between line numbers and code
    tabsize=4,                              % Size of tabs
    showstringspaces=false,                 % Don't show spaces in strings
    breaklines=true,                        % Automatic line breaking
    breakatwhitespace=true,                 % Only break lines at spaces
    escapeinside={(*@}{@*)},                % Escape to LaTeX within the code
}

\subsubsection{Sample} Randomly sample a prefix from a list of prefixes.

\begin{lstlisting}[language=Python, caption={Sample}]
def sample_prefix(prefix_list : list[str]) -> str:
    if len(prefix_list) > 0:
        prefix_index = random.randint(0, len(prefix_list)-1)
        return prefix_list[prefix_index]
    return ""
\end{lstlisting}

\subsubsection{Extract} Generate a list of potential prefixes for future inference based on the confidence of previous answers.

\begin{lstlisting}[language=Python, caption={Extract}]
def get_prefix(answers, conf_thresh, gens, prefix, prefix_level, frq:int = 5):
    confidence = confidence_criteria(answers, conf_thresh)
    prefix_list = []
    if confidence['conf'] == False:
        return prefix, prefix_level
    for i, gen in enumerate(gens[-frq:]):
        index = len(gens) - frq + i
        parts = gen.split("answer is")
        if len(parts) > 1:
            pre_gen = parts[0]
            if answers[index] == confidence['most_common']:
                parts = re.split(r'(\.\s|\.\n|\n)', pre_gen)
                sentences = []
                for i in range(0, len(parts) - 1, 2):
                    sentences.append(parts[i] + parts[i + 1])
                sentence = ""
                for j in range(prefix_level + 1):
                    if j < len(sentences) - 1:
                        sentence += sentences[j]
                        # sentence += ". "
                prefix_list.append(sentence)
    prefix_level += 1
    return prefix_list, prefix_level
\end{lstlisting}

\subsubsection{Main Inference} A class to perform path-consistency inference using a model. It recursively samples and integrates answers based on confidence thresholds, ultimately returning a final answer.

\begin{lstlisting}[language=Python, caption={Inference}]
def inference(self, prompt: str, **kwargs):
    answers = []
    prefix_level = 0
    prefix_list = []
    reasoning = []
    times = []
    generations = []
    for branch_id in range(self.max_branch): 
        prefix = sample_prefix(prefix_list)
        prompt_plus_prefix = prompt + prefix
        start_time = time.time()
        generation = self.model.completion_function(prompt_plus_prefix,**kwargs)
        end_time = time.time()
        times.append(end_time - start_time)
        reasoning.append(prefix + generation)
        answer = extract_answer(generation, self.ans_type)
        answers.append(answer)
        generations.append(generation)
        if (branch_id + 1) % (self.max_branch / (self.max_level + 1)) == 0:
            prefix_list, prefix_level = get_prefix(answers, self.confidence_thres, reasoning, prefix_list, prefix_level)   
    final_answer = integrate_answer(answers)
    info = {'answer' : final_answer, 'answers' : answers, 'latency' : times, 'generations' : generations}
    return info
\end{lstlisting}

\subsubsection{Wrap the Model}
Abstract base class for completion models defines the interface for completion models. All subclasses must implement the ``completion\_function'' method. Concrete implementation of CompletionModel using the Llama model uses a text generation model (like Llama) to generate completions based on a given prompt.

\begin{lstlisting}[language=Python, caption={Inference}]
class CompletionModel(ABC):
    @abstractmethod
    def completion_function(self, prompt: str) -> str:
        pass

class LlamaModel(CompletionModel):
    def __init__(self, generator):
        self.generator = generator

    def completion_function(self, prompt: str, **kwargs) -> str:
        prompt_list = [prompt]

        results = self.generator.text_completion(
            prompt_list,
            **kwargs
        )
        return results[-1]['generation']
\end{lstlisting}

\newpage
\subsection{Full Prompts}

\begin{table}[hbtp]
    \caption{Few-shot exemplars for arithmetic reasoning.}
    \label{tab:mathprompt}
    \centering
    % \begin{tabular}{@{}l@{}}
    \begin{tabularx}{\columnwidth}{@{}X@{}}
        \toprule
        \textbf{Q:} There are 15 trees in the grove. Grove workers will plant trees in the grove today. After they are done, there will be 21 trees. How many trees did the grove workers plant today? \\
        \textbf{A:} We start with 15 trees. Later we have 21 trees. The difference must be the number of trees they planted. So, they must have planted 21 - 15 = 6 trees. The answer is 6. \\ \\
        \textbf{Q:} If there are 3 cars in the parking lot and 2 more cars arrive, how many cars are in the parking lot? \\
        \textbf{A:} There are 3 cars in the parking lot already. 2 more arrive. Now there are 3 + 2 = 5 cars. The answer is 5. \\\\
        \textbf{Q:} Leah had 32 chocolates and her sister had 42. If they ate 35, how many pieces do they have left in total? \\
        \textbf{A:} Leah had 32 chocolates and Leah’s sister had 42. That means there were originally 32 + 42 = 74 chocolates. 35 have been eaten. So in total they still have 74 - 35 = 39 chocolates. The answer is 39. \\\\
        \textbf{Q:} Jason had 20 lollipops. He gave Denny some lollipops. Now Jason has 12 lollipops. How many lollipops did Jason give to Denny? \\
        \textbf{A:} Jason had 20 lollipops. Since he only has 12 now, he must have given the rest to Denny. The number of lollipops he has given to Denny must have been 20 - 12 = 8 lollipops. The answer is 8. \\\\
        \textbf{Q:} Shawn has five toys. For Christmas, he got two toys each from his mom and dad. How many toys does he have now? \\
        \textbf{A:} He has 5 toys. He got 2 from mom, so after that he has 5 + 2 = 7 toys. Then he got 2 more from dad, so in total he has 7 + 2 = 9 toys. The answer is 9. \\\\
        \textbf{Q:} There were nine computers in the server room. Five more computers were installed each day, from monday to thursday. How many computers are now in the server room? \\
        \textbf{A:} There are 4 days from monday to thursday. 5 computers were added each day. That means in total 4 * 5 = 20 computers were added. There were 9 computers in the beginning, so now there are 9 + 20 = 29 computers. The answer is 29. \\\\
        \textbf{Q:} Michael had 58 golf balls. On tuesday, he lost 23 golf balls. On wednesday, he lost 2 more. How many golf balls did he have at the end of wednesday? \\
        \textbf{A:} Michael initially had 58 balls. He lost 23 on Tuesday, so after that he has 58 - 23 = 35 balls. On Wednesday he lost 2 more so now he has 35 - 2 = 33 balls. The answer is 33. \\\\
        \textbf{Q:} Olivia has \$23. She bought five bagels for \$3 each. How much money does she have left? \\
        \textbf{A:} She bought 5 bagels for \$3 each. This means she spent 5 * \$3 = \$15 on the bagels. She had \$23 in beginning, so now she has \$23 - \$15 = \$8. The answer is 8. \\
        \bottomrule
    \end{tabularx}
    % \end{tabular}
\end{table}

\begin{table}[H]
    \caption{Few-shot exemplars for StrategyQA.}
    \label{tab:reasoningprompt}
    \centering
    \begin{tabularx}{\columnwidth}{@{}X@{}}
        \toprule
        \textbf{Q:} Do hamsters provide food for any animals? \\
        \textbf{A:} Hamsters are prey animals. Prey are food for predators. Thus, hamsters provide food for some animals. The answer is yes. \\ \\
        
        \textbf{Q:} Yes or no: Hydrogen's atomic number squared exceeds number of Spice Girls? \\
        \textbf{A:} Hydrogen has an atomic number of 1. 1 squared is 1. There are 5 Spice Girls. Thus, Hydrogen's atomic number squared is less than 5. The answer is no. \\ \\
        
        \textbf{Q:} Yes or no: Would a pear sink in water? \\
        \textbf{A:} The density of a pear is about 0.6 g/cm\textsuperscript{3}, which is less than water. Objects less dense than water float. Thus, a pear would float. The answer is no. \\ \\
        
        \textbf{Q:} Yes or no: Is it common to see frost during some college commencements? \\
        \textbf{A:} College commencement ceremonies can happen in December, May, and June. December is in the winter, so there can be frost. Thus, there could be frost at some commencements. The answer is yes. \\ \\
        
        \textbf{Q:} Yes or no: Could a llama birth twice during War in Vietnam (1945-46)? \\
        \textbf{A:} The War in Vietnam was 6 months. The gestation period for a llama is 11 months, which is more than 6 months. Thus, a llama could not give birth twice during the War in Vietnam. The answer is no. \\ \\
        
        \textbf{Q:} Could Brooke Shields succeed at University of Pennsylvania? \\
        \textbf{A:} Brooke Shields went to Princeton University. Princeton University is about as academically rigorous as the University of Pennsylvania. Thus, Brooke Shields could also succeed at the University of Pennsylvania. The answer is yes. \\
        \bottomrule
    \end{tabularx}
\end{table}

\begin{table}[H]
    \caption{Few-shot exemplars for RuinNames.}
    \label{tab:reasoningprompt}
    \centering
    \begin{tabularx}{\columnwidth}{@{}X@{}}
        \toprule
        \textbf{Q:} Which of the following is a humorous edit of this artist or movie name: 'whitesnake'? \\
        \textbf{Options:} (A) whitesnape, (B) whitesnapke, (C) whitesnuake, (D) mwhitesnake \\
        \textbf{A:} The original name is "whitesnake". This is the name of an old English hard rock band. It is a compound word, formed by the words "white" and "snake". \\
        (A) "whitesnape": It is formed by the combination of "white" and "snake"; therefore, "snake" has been changed to "snape". Snape makes a reference to the fictional character Severus Snape in the Harry Potter series, so (A) is indeed a meaningful and funny edit. \\
        (B) "whitesnapke": It is formed by the combination of "white" and "snapke", but "snapke" is not an actual word; therefore, "whitesnapke" is not humorous. \\
        (C) "whitesnuake": It is formed by the combination of "white" and "snuake", but "snuake" is not an actual word; therefore, "whitesnuake" is not humorous. \\
        (D) "mwhitesnake": It is formed by the combination of "m", "white", and "snake", but the prefix "-m "seems arbitrary; therefore, "mwhitesnake" is not meaningful or humorous. \\
        Above the above, the only humorous edit is (A). So the answer is (A). \\\\
        
        \textbf{Q:} Which of the following is a humorous edit of this artist or movie name: 'one of our dinosaurs is missing'? \\
        \textbf{Options:} (A) ofne of our dinosaurs is missing, (B) one af our dinosaurs is missing, (C) one of our dinosaurs is pissing, (D) one of our dinosaur is missing \\
        \textbf{A:} The original name is "one of our dinosaurs is missing". This is the name of an old British movie. \\
        (A) "ofne of our dinosaurs is missing": Here "one of" is changed to "ofne", but the word "ofne" is not an actual word. \\
        (B) "one af our dinosaurs is missing": Here the word "of" is changed to "af", but the word "af" is not an actual word. \\
        (C) "one of our dinosaurs is pissing": Here the word "missing" is changed to "pissing", and "one of our dinosaurs is pissing" is indeed a very whimsical and mischievous edit. This change truly ruins the original title of the movie. \\
        (D) "one of our dinosaur is missing": Here the word "dinosaurs" is changed to "dinosaur", but "dinosaur" is singular but should be plural in the title; this change therefore feels arbitrary and not humorous. \\
        Above the above, the only humorous edit is (C). So the answer is (C). \\\\
        
        \textbf{Q:} Which of the following is a humorous edit of this artist or movie name: 'counting crows'? \\
        \textbf{Options:} (A) countingy crows, (B) counting cows, (C) courting crows, (D) coutnting crows \\
        \textbf{A:} The original name is "counting crows". This is the name of an American rock band. Historically, the band name comes from the British nursery rhyme "One for Sorrow", which is about counting of magpies. \\
        (A) "countingy crows": Here the word "counting" is changed to "countingy", but the word "countingy" is not an actual word. \\
        (B) "counting cows": Here the word "crows" is changed to "cows", and this is indeed a playful and meaningful edit that ruins the original name of the band. \\
        (C) "courting crows": Here the word "counting" is changed to "courting", and "courting" is an actual word; however, "courting crows" does not sound as humorous as "counting cows". \\
        (D) "coutnting crows": Here the word "counting" is changed to "coutnting", but the word "coutnting" is not an actual word. \\
        Above the above, the only humorous edit is (B). So the answer is (B). \\
        \bottomrule
    \end{tabularx}
\end{table}

\begin{table}[H]
    \caption{Few-shot exemplars for SalientTranslation.}
    \label{tab:errordetection}
    \centering
    \small
    \begin{tabularx}{\columnwidth}{@{}X@{}}
        \toprule
        \textbf{Q:} The following translations from German to English contain a particular error. That error will be one of the following types: Named Entities: An entity (names, places, locations, etc.) is changed to a different entity. Numerical Values: Numerical values (ordinals or cardinals), dates, and/or units are changed. Modifiers or Adjectives: The modifiers and adjectives pertaining to a noun are changed. Negation or Antonyms: Introduce or remove a negation or change comparatives to their antonyms. Facts: Trivial factual errors not pertaining to the above classes are introduced in the translations. Dropped Content: A significant clause in the translation is removed. Please identify that error. \\
        \textbf{Source:} In der Liste der Baudenkmale in Lenzen (Elbe) sind alle Baudenkmale der brandenburgischen Stadt Lenzen (Elbe) und ihrer Ortsteile aufgelistet. \\
        \textbf{Translation:} In the list of architectural monuments in Lenzen all architectural monuments of the Brandenburg city of Lenzen and its districts are listed. \\
        \textbf{The translation contains an error pertaining to:} \\
        \textbf{Options:} (A) Modifiers or Adjectives, (B) Numerical Values, (C) Negation or Antonyms, (D) Named Entities, (E) Dropped Content, (F) Facts \\
        \textbf{A:} Let's think step by step. We solve this question by first translating the source sentence to English and then by comparing our translation with the provided translation. According to Google Translate, the correct translation of the source sentence from German to English is "The list of monuments in Lenzen (Elbe) includes all the monuments in the Brandenburg town of Lenzen (Elbe) and its districts." On the other hand, the provided translation is "In the list of architectural monuments in Lenzen all architectural monuments of the Brandenburg city of Lenzen and its districts are listed." Note that Lenzen (Elbe) is changed to Lenzen in the original translation; so, there is a named entity error. Because an entity in the original source sentence is changed to a different entity in the translation, the translation contains an error pertaining to Named Entities. So the answer is (D). \\ \\
        
        \textbf{Q:} The following translations from German to English contain a particular error. That error will be one of the following types: Named Entities: An entity (names, places, locations, etc.) is changed to a different entity. Numerical Values: Numerical values (ordinals or cardinals), dates, and/or units are changed. Modifiers or Adjectives: The modifiers and adjectives pertaining to a noun are changed. Negation or Antonyms: Introduce or remove a negation or change comparatives to their antonyms. Facts: Trivial factual errors not pertaining to the above classes are introduced in the translations. Dropped Content: A significant clause in the translation is removed. Please identify that error. \\
        \textbf{Source:} Auf dieser Seite sind die Baudenkmaler der oberbayerischen Groben Kreisstadt Landsberg am Lech zusammengestellt. \\
        \textbf{Translation:} On this page are compiled the architectural monuments of the town of Landsberg am Lech. \\
        \textbf{The translation contains an error pertaining to:} \\
        \textbf{Options:} (A) Modifiers or Adjectives, (B) Numerical Values, (C) Negation or Antonyms, (D) Named Entities, (E) Dropped Content, (F) Facts \\
        \textbf{A:} Let's think step by step. We solve this question by first translating the source sentence to English and then by comparing our translation with the provided translation. According to Google Translate, the correct translation of the source sentence from German to English is "The monuments of the Upper Bavarian district town of Landsberg am Lech are compiled on this page." On the other hand, the provided translation is "On this page are compiled the architectural monuments of the town of Landsberg am Lech." Note that an important detail about the location of Landsberg am Lech is omitted in the original translation: The translation should have said "Upper Bavarian district town of Landsberg am Lech". Because a significant clause in the translation was removed, the translation contains an error pertaining to Dropped Content. So the answer is (E). \\ \\
        
        \textbf{Q:} The following translations from German to English contain a particular error. That error will be one of the following types: Named Entities: An entity (names, places, locations, etc.) is changed to a different entity. Numerical Values: Numerical values (ordinals or cardinals), dates, and/or units are changed. Modifiers or Adjectives: The modifiers and adjectives pertaining to a noun are changed. Negation or Antonyms: Introduce or remove a negation or change comparatives to their antonyms. Facts: Trivial factual errors not pertaining to the above classes are introduced in the translations. Dropped Content: A significant clause in the translation is removed. Please identify that error. \\
        \textbf{Source:} Leba ist eine Kleinstadt und ein Badeort im Powiat Leborski der polnischen Woiwodschaft Pommern. \\
        \textbf{Translation:} Eba is not a small town and seaside resort in the Powiat Leborski county of the Pomeranian Voivodeship of Poland. \\
        \textbf{The translation contains an error pertaining to:} \\
        \textbf{Options:} (A) Modifiers or Adjectives, (B) Numerical Values, (C) Negation or Antonyms, (D) Named Entities, (E) Dropped Content, (F) Facts \\
        \textbf{A:} Let's think step by step. We solve this question by first translating the source sentence to English and then by comparing our translation with the provided translation. According to Google Translate, the correct translation of the source sentence from German to English is "Łeba is a small town and seaside resort in the Powiat Leborski of the Polish Pomeranian Voivodeship." On the other hand, the provided translation is "Leba is not a small town and seaside resort in the Powiat Leborski county of the Pomeranian Voivodeship of Poland." Note that the provided sentence says, "Leba is not a small town ..." However, the translation should have been "Leba is a small town ..." Because a negation is introduced at the beginning of the sentence and has fundamentally changed the meaning of the original source, the translation contains an error pertaining to Negation or Antonyms. So the answer is (C).\\
        \bottomrule
    \end{tabularx}
\end{table}

\begin{table}[H]
    \caption{Few-shot exemplars for BooleanExpressions.}
    \label{tab:booleanexpressions}
    \centering
    \begin{tabularx}{\columnwidth}{@{}X@{}}
        \toprule
        \textbf{Q:} not ( ( not not True ) ) is \\
        \textbf{A:} Let's think step by step. \\
        Remember that (i) expressions inside brackets are always evaluated first and that (ii) the order of operations from highest priority to lowest priority is "not", "and", "or", respectively. \\
        We first simplify this expression "Z" as follows: "Z = not ( ( not not True ) ) = not ( ( A ) )" where "A = not not True". \\
        Let's evaluate A: A = not not True = not (not True) = not False = True. \\
        Plugging in A, we get: Z = not ( ( A ) ) = not ( ( True ) ) = not True = False. So the answer is False. \\ \\
        
        \textbf{Q:} True and False and not True and True is \\
        \textbf{A:} Let's think step by step. \\
        Remember that (i) expressions inside brackets are always evaluated first and that (ii) the order of operations from highest priority to lowest priority is "not", "and", "or", respectively. \\
        We first simplify this expression "Z" as follows: "Z = True and False and not True and True = A and B" where "A = True and False" and "B = not True and True". \\
        Let's evaluate A: A = True and False = False. \\
        Let's evaluate B: B = not True and True = not (True and True) = not (True) = False. \\
        Plugging in A and B, we get: Z = A and B = False and False = False. So the answer is False. \\ \\
        
        \textbf{Q:} not not ( not ( False ) ) is \\
        \textbf{A:} Let's think step by step. \\
        Remember that (i) expressions inside brackets are always evaluated first and that (ii) the order of operations from highest priority to lowest priority is "not", "and", "or", respectively. \\
        We first simplify this expression "Z" as follows: "Z = not not ( not ( False ) ) = not not ( A )" where "A = not ( False )". \\
        Let's evaluate A: A = not ( False ) = not False = True. \\
        Plugging in A, we get: Z = not not ( A ) = not not (True) = not not False = True. So the answer is True. \\
        \bottomrule
    \end{tabularx}
\end{table}

\begin{table}[H]
    \caption{Few-shot exemplars for TrackingShuffledObject.}
    \label{tab:finalpositions}
    \centering
    \begin{tabularx}{\columnwidth}{@{}X@{}}
        \toprule
        \textbf{Q:} Alice, Bob, and Claire are playing a game. At the start of the game, they are each holding a ball: Alice has a yellow ball, Bob has a blue ball, and Claire has a pink ball. As the game progresses, pairs of players trade balls. First, Claire and Alice swap balls. Then, Alice and Bob swap balls. Finally, Claire and Bob swap balls. At the end of the game, Bob has the \\
        \textbf{Options:} (A) yellow ball, (B) blue ball, (C) pink ball \\
        \textbf{A:} Let's think step by step. \\
        (0) At the start: Alice: yellow, Bob: blue, Claire: pink. \\
        (1) Claire and Alice swap balls: Alice: pink, Bob: blue, Claire: yellow. \\
        (2) Alice and Bob swap balls: Alice: blue, Bob: pink, Claire: yellow. \\
        (3) Claire and Bob swap balls: Alice: blue, Bob: yellow, Claire: pink. \\
        At the end of the game, Bob has the yellow ball. So the answer is (A). \\ \\
        
        \textbf{Q:} Alice, Bob, and Claire are playing a game. At the start of the game, they are each holding a ball: Alice has a white ball, Bob has a purple ball, and Claire has a pink ball. As the game progresses, pairs of players trade balls. First, Bob and Alice swap balls. Then, Bob and Claire swap balls. Finally, Bob and Alice swap balls. At the end of the game, Alice has the \\
        \textbf{Options:} (A) white ball, (B) purple ball, (C) pink ball \\
        \textbf{A:} Let's think step by step. \\
        (0) At the start: Alice: white, Bob: purple, Claire: pink. \\
        (1) Bob and Alice swap balls: Alice: purple, Bob: white, Claire: pink. \\
        (2) Bob and Claire swap balls: Alice: purple, Bob: pink, Claire: white. \\
        (3) Bob and Alice swap balls: Alice: pink, Bob: purple, Claire: white. \\
        At the end of the game, Alice has the pink ball. So the answer is (C). \\ \\
        
        \textbf{Q:} Alice, Bob, and Claire are dancers at a square dance. At the start of a song, they each have a partner: Alice is dancing with Lola, Bob is dancing with Rodrigo, and Claire is dancing with Patrick. Throughout the song, the dancers often trade partners. First, Alice and Bob switch partners. Then, Claire and Bob switch partners. Finally, Bob and Alice switch partners. At the end of the dance, Alice is dancing with \\
        \textbf{Options:} (A) Lola, (B) Rodrigo, (C) Patrick \\
        \textbf{A:} Let's think step by step. \\
        (0) At the start: Alice: Lola, Bob: Rodrigo, Claire: Patrick. \\
        (1) Alice and Bob switch partners: Alice: Rodrigo, Bob: Lola, Claire: Patrick. \\
        (2) Claire and Bob switch partners: Alice: Rodrigo, Bob: Patrick, Claire: Lola. \\
        (3) Bob and Alice switch partners: Alice: Patrick, Bob: Rodrigo, Claire: Lola. \\
        At the end of the dance, Alice is dancing with Patrick. So the answer is (C). \\
        \bottomrule
    \end{tabularx}
\end{table}

\begin{table}[H]
    \caption{Few-shot exemplars for LogicalDeduction.}
    \label{tab:logicaldeduction}
    \centering
    \begin{tabularx}{\columnwidth}{@{}X@{}}
        \toprule
        \textbf{Q:} The following paragraphs each describe a set of three objects arranged in a fixed order. The statements are logically consistent within each paragraph. In a golf tournament, there were three golfers: Amy, Eli, and Eve. Eve finished above Amy. Eli finished below Amy. \\
        \textbf{Options:} (A) Amy finished last, (B) Eli finished last, (C) Eve finished last \\
        \textbf{A:} Let's think step by step. \\
        (1) Eve finished above Amy: "(above) ? Eve ? Amy ? (below)". \\
        (2) Eli finished below Amy: "(above) ? Amy ? Eli ? (below)". \\
        (3) Combining (1) and (2) we get the following ordering: "(above) Eve Amy Eli (below)". \\
        According to this ordering, the person who finished last (the one at the bottom of this list) is Eli. \\
        Eli finished last. So the answer is (B). \\ \\
        
        \textbf{Q:} The following paragraphs each describe a set of three objects arranged in a fixed order. The statements are logically consistent within each paragraph. On a shelf, there are three books: a white book, a green book, and an orange book. The green book is to the right of the white book. The orange book is the rightmost. \\
        \textbf{Options:} (A) The white book is the leftmost, (B) The green book is the leftmost, (C) The orange book is the leftmost \\
        \textbf{A:} Let's think step by step. \\
        (1) The green book is to the right of the white book: "(left) ? white ? green ? (right)". \\
        (2) The orange book is the rightmost: "(left) ? white ? green orange (right)". \\
        (3) Combining (1) and (2) we get the following ordering: "(left) white green orange (right)". \\
        According to this ordering, the leftmost book is the white book. \\
        The white book is the leftmost. So the answer is (A). \\ \\
        
        \textbf{Q:} The following paragraphs each describe a set of three objects arranged in a fixed order. The statements are logically consistent within each paragraph. On a shelf, there are three books: a red book, a gray book, and a white book. The white book is to the left of the gray book. The red book is the second from the left. \\
        \textbf{Options:} (A) The red book is the leftmost, (B) The gray book is the leftmost, (C) The white book is the leftmost \\
        \textbf{A:} Let's think step by step. \\
        (1) The white book is to the left of the gray book: "(left) ? white ? gray ? (right)". \\
        (2) The red book is the second from the left: "(left) ? white red gray ? (right)". \\
        (3) Combining (1) and (2) we get the following ordering: "(left) white red gray (right)". \\
        According to this ordering, the leftmost book is the white book. \\
        The white book is the leftmost. So the answer is (C). \\
        \bottomrule
    \end{tabularx}
\end{table}

\end{document}